\documentclass[10pt,twocolumn,letterpaper]{article}

\usepackage[pagenumbers]{cvpr} 

\usepackage{times}
\usepackage{epsfig}
\usepackage{graphicx}
\usepackage{amsmath}
\usepackage{amssymb}

\usepackage{booktabs}
\usepackage{tabulary}
\usepackage[dvipsnames]{xcolor}
\usepackage{soul}
\usepackage{xspace}\xspace
\usepackage{adjustbox}
\usepackage{array}

\usepackage{dblfloatfix}
\usepackage{transparent}

\usepackage{algorithm}
\usepackage{listings}

\usepackage{etoolbox}
\makeatletter
\AfterEndEnvironment{algorithm}{\let\@algcomment\relax}
\AtEndEnvironment{algorithm}{\kern2pt\hrule\relax\vskip3pt\@algcomment}
\let\@algcomment\relax
\newcommand\algcomment[1]{\def\@algcomment{\footnotesize#1}}
\renewcommand\fs@ruled{\def\@fs@cfont{\bfseries}\let\@fs@capt\floatc@ruled
	\def\@fs@pre{\hrule height.8pt depth0pt \kern2pt}%
	\def\@fs@post{}%
	\def\@fs@mid{\kern2pt\hrule\kern2pt}%
	\let\@fs@iftopcapt\iftrue}
\makeatother

\usepackage{color}
\definecolor{mybar}{rgb}{1.0, 0.0, 1.0}
\definecolor{myorange}{RGB}{255, 145, 1.0}
\definecolor{mypink}{RGB}{254, 0, 254}

\usepackage{multirow}
\usepackage{makecell}
\usepackage{threeparttable}
\usepackage[pagebackref,breaklinks,colorlinks]{hyperref}

\usepackage[capitalize]{cleveref}
\crefname{section}{Sec.}{Secs.}
\Crefname{section}{Section}{Sections}
\Crefname{table}{Table}{Tables}
\crefname{table}{Tab.}{Tabs.}

\def\cvprPaperID{****} 
\def\confName{****}
\def\confYear{2023}

\begin{document}


\title{EHSNet: End-to-End Holistic Learning Network for Large-Size 

Remote Sensing Image Semantic Segmentation}


\author{Wei Chen\qquad Yansheng Li\thanks{corresponding author.}\qquad Bo Dang \qquad Yongjun Zhang \\
Wuhan University, Wuhan, China\\
{\tt\small \{weichenrs, yansheng.li, bodang, zhangyj\}@whu.edu.cn}
}

\maketitle

\begin{abstract}
    This paper presents \textbf{EHSNet}, a new end-to-end segmentation network designed for the holistic learning of large-size remote sensing image semantic segmentation (LRISS). Large-size remote sensing images (LRIs) can lead to GPU memory exhaustion due to their extremely large size, which has been handled in previous works through either global-local fusion or multi-stage refinement, both of which are limited in their ability to fully exploit the abundant information available in LRIs. Unlike them, EHSNet features three memory-friendly modules to utilize the characteristics of LRIs: a long-range dependency module to develop long-range spatial context, an efficient cross-correlation module to build holistic contextual relationships, and a boundary-aware enhancement module to preserve complete object boundaries. Moreover, EHSNet manages to process holistic LRISS with the aid of memory offloading. To the best of our knowledge, EHSNet is the first method capable of performing holistic LRISS. To make matters better, EHSNet outperforms previous state-of-the-art competitors by a significant margin of +5.65 mIoU on FBP and +4.28 mIoU on Inria Aerial, demonstrating its effectiveness. We hope that EHSNet will provide a new perspective for LRISS. The code and models will be made publicly available.
\end{abstract}

\begin{figure}[t]
  \centering
   \includegraphics[width=\linewidth]{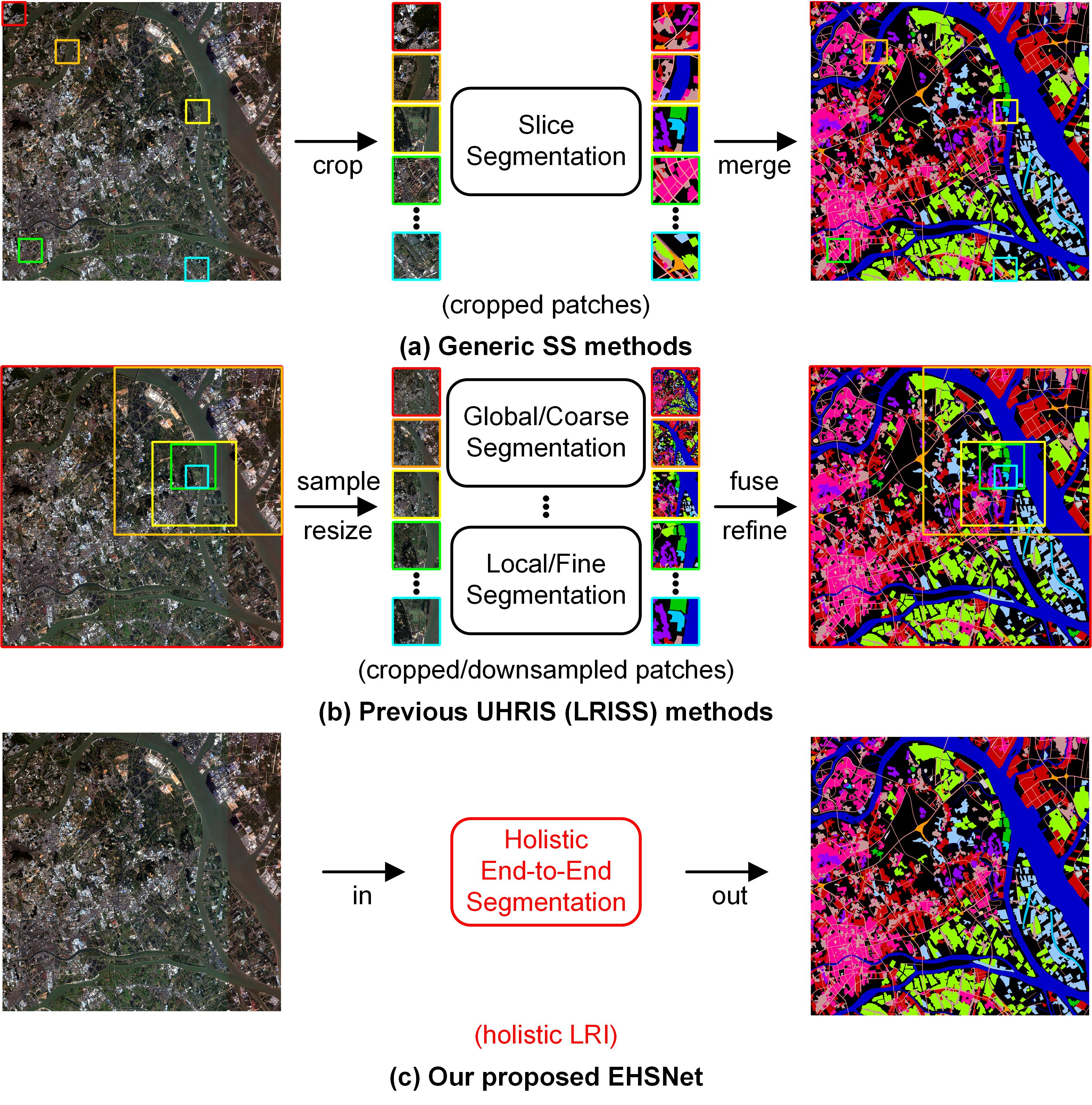}

   \caption{\textbf{Comparison of LRISS pipelines:} (a) Generic SS methods can only process cropped patches from LRIs and merge the results into a whole output. (b) Previous LRISS methods often train multiple models with samples of different sizes and conduct global-to-local fusion or coarse-to-fine refinement on multi-stage predictions. (c) Our EHSNet can perform LRISS by end-to-end holistic learning without merging, fusion, or refinement.}
   \label{fig:1}
\end{figure}

\section{Introduction}
\label{sec:intro}

With the advancements in photography and sensor technologies, large-size remote sensing images (LRIs) containing several tens of millions of pixels can now be accessed by humans, consisting of aerial images \cite{inria}, satellite images \cite{fbp}, and others. The effective use of LRIs will benefit a wild range of technologies such as building extraction \cite{building2021} and land-cover classification \cite{mfvnet}, which will facilitate further applications like urban planning \cite{rsapp1} and emergency response \cite{emergency2019}.

Semantic segmentation (SS) is one of the most fundamental vision tasks that provides a basic but essential understanding of images. From the earliest fully convolutional networks \cite{fcn} to recent Transformer-based models \cite{swin, segformer, setr, strudel2021segmenter, liu2022swin}, SS has witnessed an explosion of deep learning-based techniques with continuously growing capability and capacity \cite{pspnet, deeplabv3p, zhao2018psanet, xiao2018upernet, hrnet, yuan2020ocrnet, kirillov2020pointrend}. However, despite the significant progress made in generic SS, large-size remote sensing image semantic segmentation (LRISS) lags behind, with only preliminary explorations having been done \cite{glnet, fctl, magnet, guo2022isdnet, mfvnet}. Therefore, the question arises: \textbf{what sets LRISS apart from generic SS?} In this paper, we attempt to answer this question from several perspectives.

\begin{figure}[t]
  \centering
   \includegraphics[width=\linewidth]{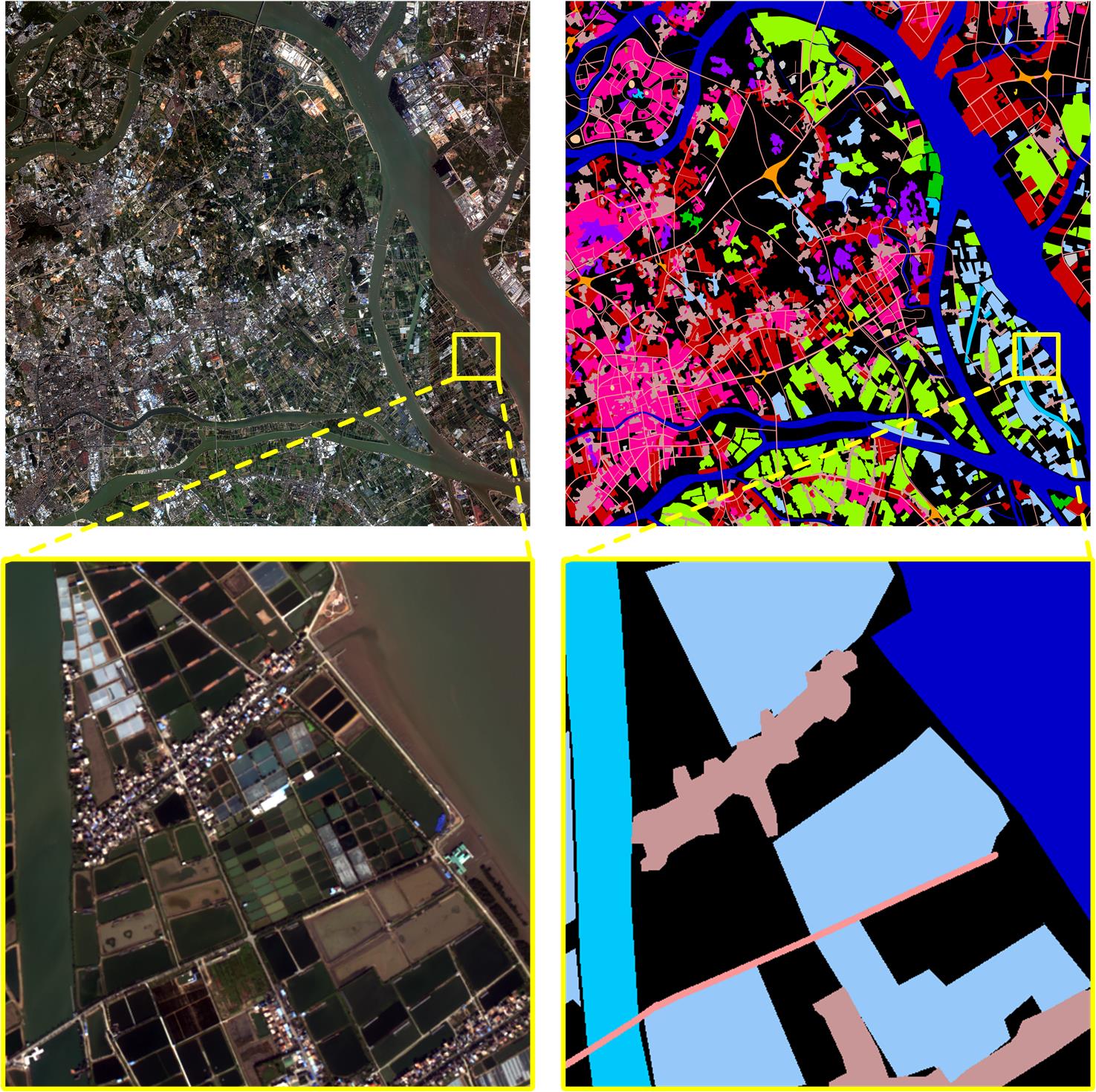}
    \caption{\textbf{The necessity of holistic LRISS:} The lower two images demonstrate that using cropped slices for LRISS can lead to ambiguous predictions. Even humans cannot always definitively distinguish the {\color{cyan}{pond}} on the left side from the {\color{blue}{river}} on the right side. However, with the help of the holistic perspective provided by the LRI, they can be distinguished with complete structure and rich spatial context.
    }
   \label{fig:2}
\end{figure}

\textbf{(i)} LRIs offer a comprehensive view of the scene, providing both macroscopic structures and microcosmic details, unlike generic-sized images that have limited contexts. As shown in \cref{fig:2}, the global view provides complete boundaries of the LRI, while the local view helps to bring fine information for segmenting specific regions. A combination of the two views leads to accurate segmentation results, which can be seen from the results in \cref{fig:6}. This approach is similar to human vision \cite{humanvision1, humanvision2, humanvision}.

\textbf{(ii)} LRISS requires not only assigning a category to each pixel in the image but also seeking high-level semantics from a broader perspective. Taking satellite images as examples, instead of focusing on object-level categories such as cars, buildings, and trees, LRISS needs models to capture long-range relationships among various regions for SS with complex categories. For example, groups of houses and trees in LRISS are identified as an "urban area" \cite{castillo2021semi, gongpeng}. However, pixel-by-pixel SS may lead to confusing results under such circumstances.

\textbf{(iii)} LRIs can contain millions or even billions of pixels, posing a huge challenge in terms of memory and computation when performing holistic segmentation. Previous LRISS methods \cite{glnet, fctl, magnet, guo2022isdnet, mfvnet} all fail to fit whole LRIs into the model, adopting a suboptimal approach to fuse or refine the results of patches with less information.

Considering these, we present a simple but effective paradigm named EHSNet for LRISS. Unlike previous LRISS methods \cite{glnet, fctl, magnet, guo2022isdnet, mfvnet}, our EHSNet can holistically process LRISS by utilizing memory offloading. With the aid of three specially-designed memory-friendly modules, the characteristics of LRIs are thoroughly exploited. Extensive experiments on two typical LRISS datasets \cite{fbp, inria} demonstrate the superiority of EHSNet over previous state-of-the-arts (SOTAs).

The main contributions are summarized as follows:
\begin{itemize}
\item[$\bullet$] We propose a novel network called EHSNet, with three memory-friendly modules to exploit information better in end-to-end holistic LRISS. To the best of our knowledge, this is the first attempt at conducting LRISS holistically.
\item[$\bullet$] We introduce a long-range dependency (LRD) module and an efficient cross-correlation (ECR) module to incorporate long-range spatial context and build holistic contextual relationships, respectively.
\item[$\bullet$] Extensive experiments and analyses on two typical LRISS datasets demonstrate the efficacy of EHSNet and its significant superiority over previous SOTA approaches.
\end{itemize}

\section{Related work}
\label{sec:relate}
\subsection{Generic semantic segmentation}
Since the introduction of Fully Convolutional Networks \cite{fcn}, generic SS methods have been leveraging multi-scale feature fusion to overcome the limitations of local receptive fields by convolutional layers and capture contextual information at multiple scales \cite{deeplabv3p, pspnet, hrnet}. For example, DeepLabV3+ \cite{deeplabv3p} uses several parallel atrous convolutions with different rates, while PSPNet \cite{pspnet} performs pooling operations at different grid scales. HRNet \cite{hrnet} connects high-to-low resolution convolutions in parallel and performs multi-scale fusion across parallel convolutions to generate strong and spatially precise high-resolution representations. Recently, the Vision Transformers \cite{vit} have made significant breakthroughs in the computer vision domain by leveraging self-attention mechanisms to capture the global spatial context in generic SS tasks \cite{swin, liu2022swin, segformer, setr}. However, these methods designed for generic images do not perform well in the context of LRISS. In contrast, our EHSNet achieves better performance by leveraging three specifically designed modules to fully utilize the characteristics of the LRIs.

\begin{figure*}
  \centering
   \includegraphics[width=\linewidth]{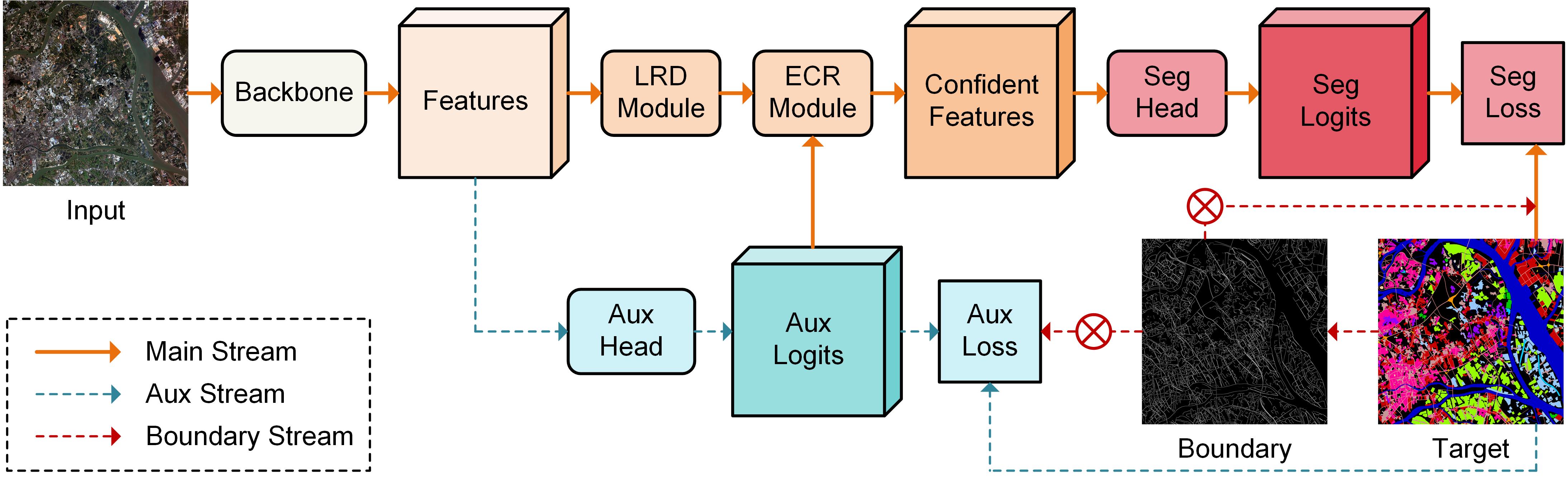}
   \caption{\textbf{Overview of the proposed EHSNet.} It consists of a backbone network, a LRD module, an ECR module, a BAE module, an auxiliary segmentation head, and a main segmentation head.}
   \label{fig:3}
\end{figure*}

\subsection{Ultra-high resolution image segmentation}
Ultra-high resolution image segmentation (UHRIS) has been a topic of research in the computer vision domain for several years and is closely related to LRISS. From the earliest GLNet \cite{glnet} to the most recent ISDNet \cite{guo2022isdnet}, many approaches have been proposed to address this challenge. It is widely believed that cropping and downsampling can harm the performance of UHRIS. To overcome this limitation, GLNet \cite{glnet} proposed a two-stream network that separately processes the downsampled global image and cropped local patches and a feature-sharing module that shares the concatenated local and global features in both streams. CascadePSP \cite{caspsp} treats UHRIS as a refinement task and uses a global step to refine the entire image, providing sufficient image contexts for the subsequent local step to perform full-resolution high-quality refinement. Following similar patterns of the two, other approaches either train multiple models with different-sized samples and conduct global-to-local fusion on the predictions \cite{tsmta, wu2020ppn, fctl, guo2022isdnet, wiconet, mfvnet} or adopt a refinement scheme that performs coarse-to-fine refinement on multi-stage predictions \cite{li2017notallrefine, lin2017refinenet, yuan2020segfix, kirillov2020pointrend, magnet, shen2022highquality} for better segmentation results. Different from previous works, our EHSNet performs LRISS through end-to-end holistic learning without merging, fusion, or refinement. Furthermore, EHSNet outperforms previous state-of-the-art methods in this regard.

\subsection{Holistic image segmentation}
It is crucial to fully utilize the rich information in LRIs for LRISS. However, previous approaches \cite{glnet, tsmta, wu2020ppn, fctl, guo2022isdnet, wiconet, mfvnet, magnet, shen2022highquality} can only handle cropped or downsampled patches and require fusion or refinement. Our experiments with their codes reveal that even ISDNet \cite{guo2022isdnet}, which has the largest operable size (2000$\times$2000 pixels), falls short of meeting the requirements of larger LRIs such as images in FBP (6800$\times$7200 pixels) or Inria Aerial (5000$\times$5000 pixels) dataset. Fortunately, the recent success of natural language processing models \cite{gpt} brings us their training strategies \cite{rasley2020deepspeed, jiang2020byteps, lms}. LMS \cite{lms}, as a representative one, enables the successful training of deep learning models that would otherwise exhaust GPU memory and result in out-of-memory errors. LMS manages GPU memory over-subscription by temporarily swapping tensors to host memory when not needed, which allows LRIs to fit into the model. With sufficient host memory, the limited GPU memory is no longer a constraint for LRISS. LMS thus makes it possible for EHSNet to process holistic segmentation.

\section{Method}
\label{sec:method}
\subsection{Overview}
An overview of EHSNet is presented in \cref{fig:3}. It comprises a backbone network, a LRD module, an ECR module, a boundary-aware enhancement (BAE) module, an auxiliary segmentation head, and a main segmentation head, which will be discussed in detail in the following sections.

\subsection{Long-range dependency module}

The LRI usually consists of geospatial objects of different sizes. Some objects such as rivers may span across the entire LRI, as shown in the upper two images in \cref{fig:2}. It is essential to combine together the local view and the global view, to preserve segmentation results with fine details and complete boundaries simultaneously. Although the stack of convolutions in the networks can be used for exploiting global information, it still lacks a complete field-of-view due to the small effective receptive field \cite{ding2022replknet}. 

\begin{figure}[t]
  \centering
   \includegraphics[width=\linewidth]{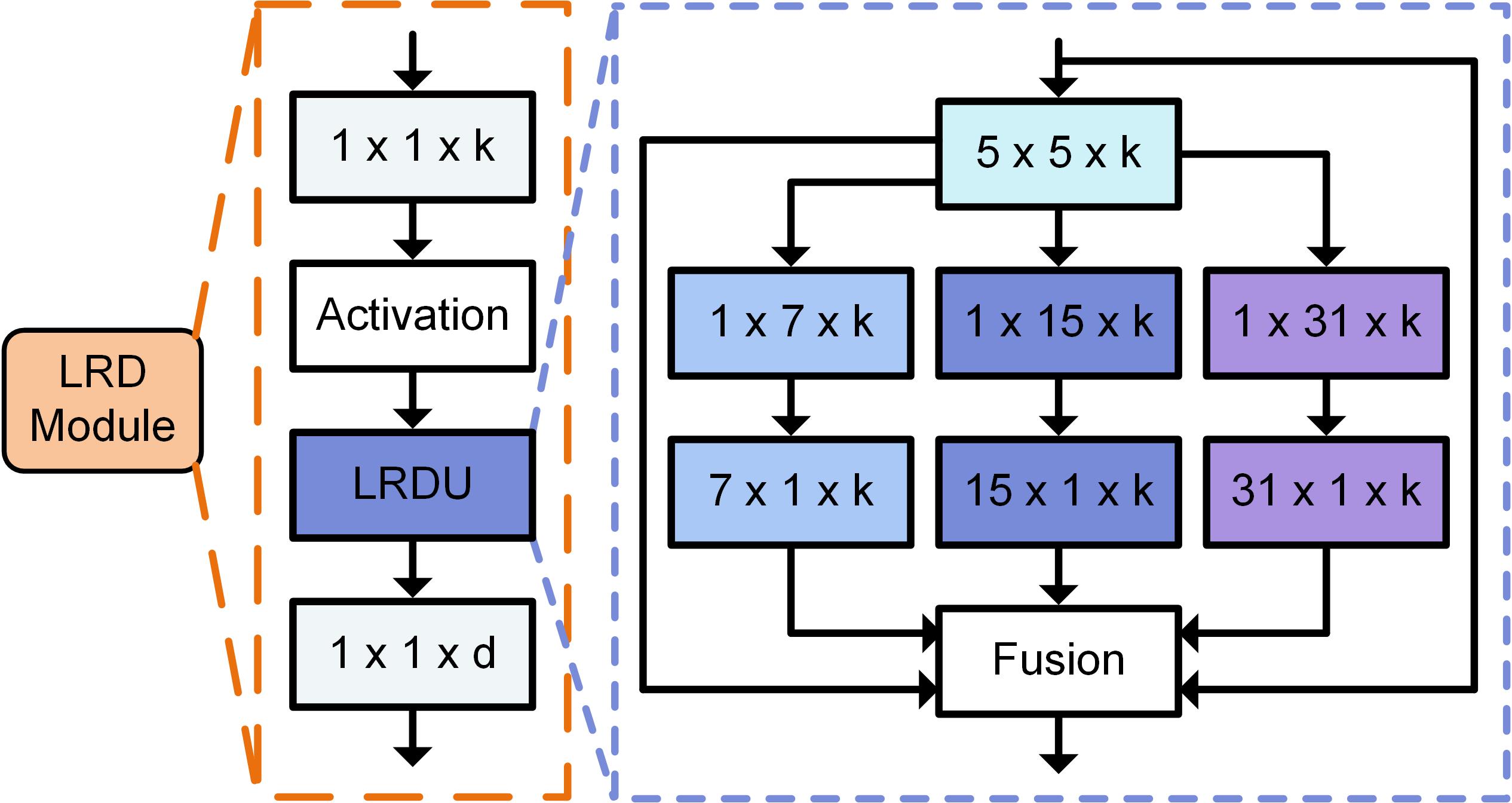}
   \caption{\textbf{Details of the proposed long-range dependency module.} With three parallel convolution kernels with sizes up to 31$\times$31, it can better build the long-range dependency in the LRIs and enlarge the effective receptive field of the EHSNet.}
   \label{fig:4}
\end{figure}

Thus, we design the LRD module in EHSNet. For LRIs, the effective receptive field is quite important for capturing global structures. Therefore, we use several parallel convolutions with large kernel sizes in the LRD module. The LRD module comprises a 1$\times$1 convolution for dimension reduction, an activation function, a Long-Range Dependency Unit (LRDU), and a 1$\times$1 convolution for dimension increasing, as depicted in \cref{fig:4}. The LRDU is the core of the LRD module and includes a 5$\times$5 convolution, three parallel large-kernel depth-wise separable convolutions (7$\times$7, 15$\times$15, and 31$\times$31), and a fusion function.


The LRD module can be depicted as \cref{eq:3},
\begin{equation}
   f_{out} = \delta(f_{in}, f_{5}, f_{7}, f_{15}, f_{31}),
  \label{eq:3}
\end{equation}
where $f_{in}$ and $f_{out}$ denote the input and output feature for the LRD module, $f_{x}$ denotes the result of convolution with a kernel size of $x$, $\delta$ denotes the combination of a convolution with a kernel size of 1 and a fusion function, which is $concat$ here.

It is noted that the dilated convolutions \cite{deeplabv3p} and the deformable convolutions \cite{dai2017deformable} are also used to enlarge the effective receptive field, but they are not considered in EHSNet due to the severe grid effects and the heavy computations, respectively. Besides, large convolutional kernels have been applied to generic SS tasks to perform large convolutional kernels for larger spatial contexts in RepLKNet \cite{ding2022replknet} and LKM \cite{peng2017largekernelmatters}. Different from them, the three parallel convolutions with different kernel sizes utilized in the LRDU lead to the better building of multi-scale long-range dependency in the LRIs.

\subsection{Efficient cross-correlation module}
We argue that LRISS is not simply allocating a category to each pixel in the image, but also pursuing high-level semantics. It is necessary to take into account the relationship among pixels, regions, and categories, the combination of which forms the spatial context. For one target pixel, we believe that all relevant pixels in the LRI can contribute to it holistically. One prevailing approach to conduct this is self-attention. Pioneers \cite{wang2018nonlocal, zhao2020selfattn, yuan2020ocrnet} have explored self-attention for capturing global contexts. However, the cost of self-attention is much too heavy for LRIs, which is impossible to be conducted. 

To make self-attention feasible for LRISS, we propose the ECR module by performing attention to the hard region with low confidence. The ECR module $\theta_{ECR}$ is depicted below. Specifically, we first compute the difference between the top2 logits $P_{1st}$ and $P_{2nd}$ from the auxiliary head. Then we sort the difference by descending and select the maximum $a\%$ region, which is considered hard region $R$ with low confidence. This procedure is depicted as $\phi$ in \cref{eq:5}, 
\begin{equation}
  R = \phi(P_{1st}, P_{2nd}).
  \label{eq:5}
\end{equation}

After that, we gather features $F_{hard}$ from the hard region $R$ of the input features $F_{in}$, perform efficient attention $\theta$, and scatter the enhanced features $F_{conf}$ to the hard region to get the output features $F_{out}$, just as depicted from \cref{eq:6} to \cref{eq:8},
\begin{equation}
  F_{hard} = gather(F_{in}, R),
  \label{eq:6}
\end{equation}
\begin{equation}
  F_{conf} = \theta(F_{in}, F_{hard}),
  \label{eq:7}
\end{equation}
\begin{equation}
  F_{out} = scattar(F_{conf}, R).
  \label{eq:8}
\end{equation}

The use of efficient attention is inspired by \cite{shen2021efficientattn}. Different from generic attention, efficient attention is performed on the key and value before multiplying with the query. See the Appendix for the proof of equivalence between generic attention and efficient attention and their comparison.

\begin{figure*}
  \centering
  \includegraphics[width=0.9\linewidth]{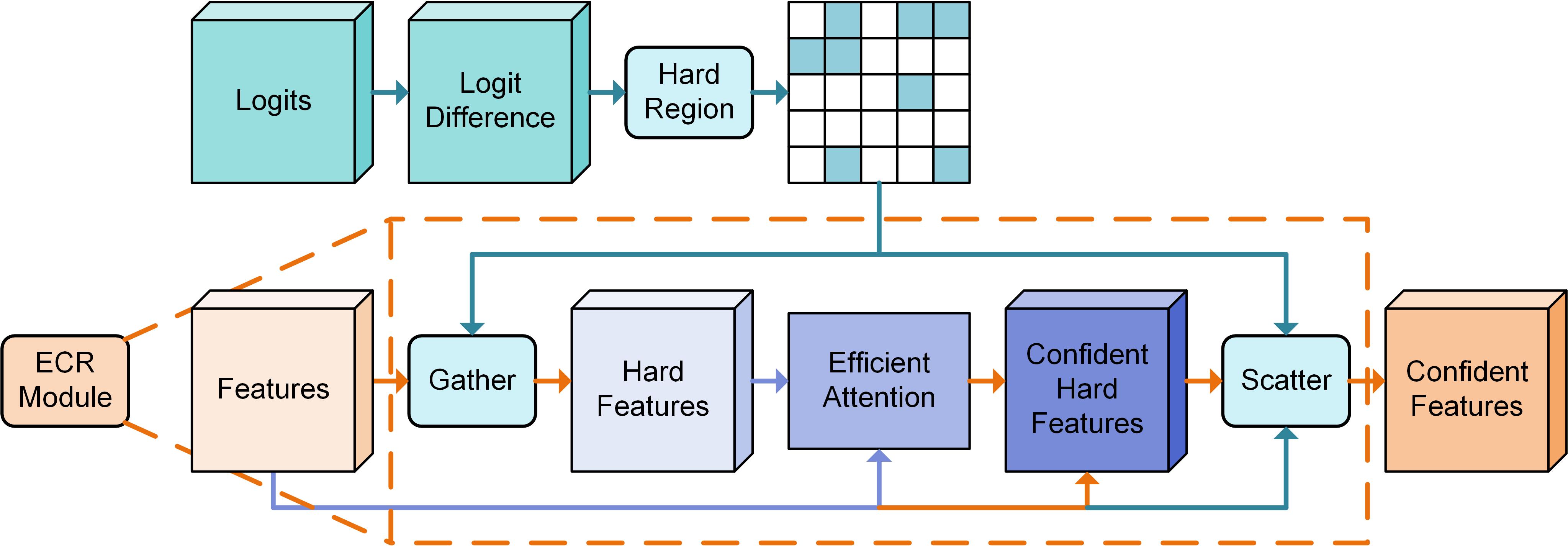}
   \caption{\textbf{Details of the proposed efficient cross-correlation module.} It performs attention to the hard region with low confidence with the aid of efficient attention, which can help to develop holistic contextual relationships between regions.}
   \label{fig:5}
\end{figure*}


\subsection{Boundary-aware enhancement module}

The boundary region is quite important for SS \cite{yuan2020segfix, kirillov2020pointrend, boundary1, boundary2, caspsp}. Typically, HQS \cite{shen2022highquality} and CascadePSP \cite{caspsp} propose effective models to refine the segmentation results on the boundary region, which works on cropped patches of the LRIs. However, these methods are much too heavy requiring too much computation and memory, which is unacceptable, especially for the holistic segmentation of LRIs. 

We believe that the boundary region is important but not necessary in the task of LRISS. Thus, we use the simplest way to enhance that in our BAE module. We naively use the Sobel operator to extract the boundaries. Different from existing methods, we perform the Sobel operator on the masks rather than the images to pursue more complete object boundaries. Then the boundaries are used as pixel-level loss weights to punish the model on the boundary region for better learning of pixel relationship on the boundary region. The BAE module can be depicted as \cref{eq:9},

\begin{equation}
  W_{bd} = I + \mathbb{I}(sobel(M)>0),
  \label{eq:9}
\end{equation}
where $I$ denotes the matrix whose elements are all 1, $M$ denotes the mask, $W_{bd}$ denotes the pixel-level loss weight for boundary enhancement, $\mathbb{I}$ is equal to 1 when $sobel(M)>0$.

The BAE module is simple but effective, which is especially useful in holistic segmentation to preserve complete boundaries for better segmentation but brings no extra computation and memory costs.

\subsection{Loss function}
EHSNet contains a backbone network with two cascaded modules, LRD and ECR. To better train the EHSNet model, we use the auxiliary segmentation loss for deep supervision, following previous works \cite{lee2015deepsupervision, pspnet}. Cross entropy loss with the proposed BAE module is used for both the main segmentation loss and the auxiliary segmentation loss, as depicted in \cref{eq:12} and \cref{eq:15}.
\begin{equation}
  \mathcal{L}_{aux} = \mathcal{L}_{CE}(P_{aux}, M, W_{bd}),
  \label{eq:12}
\end{equation}
\begin{equation}
  \mathcal{L}_{seg} = \mathcal{L}_{CE}(P_{seg}, M, W_{bd}),
  \label{eq:15}
\end{equation}
where $M$ is the corresponding mask of the input image, $P_{aux}$ and $P_{seg}$ are the outputs of the auxiliary head and the main segmentation head, $\mathcal{L}_{CE}$ is the cross entropy loss function.

The whole loss function is depicted as \cref{eq:16},
\begin{equation}
  \mathcal{L} = \mathcal{L}_{seg} + \lambda * \mathcal{L}_{aux}.
  \label{eq:16}
\end{equation}
where $\lambda$ is the weight for auxiliary loss.

\section{Experiment}
\label{sec:experiment}
\subsection{Datasets}


\textbf{FBP \cite{fbp}.} 
The FBP dataset is the largest LRISS dataset available that contains more than 5 billion labeled pixels of 150 LRIs (6800$\times$7200 pixels), annotated in a 24-category system covering artificial-constructed, agricultural, and natural classes. It possesses the advantage of rich categories, large coverage, wide distribution, and high spatial resolution, which well reflects the distributions of real-world ground objects and can benefit different land-cover-related studies. We follow the same testing sets with \cite{fbp} and randomly split the rest images into training and validation with 90 and 30 images, respectively. 

\textbf{Inria Aerial \cite{inria}}
The Inria Aerial dataset covers diverse urban landscapes, ranging from dense metropolitan districts to alpine resorts. It provides 180 LRIs (5000$\times$5000 pixels) from five cities and dense annotations with a binary mask for building and non-building areas. Following the protocols of \cite{glnet, fctl, guo2022isdnet}, we split images into training, validation, and test set with 126, 27, and 27 images, respectively.

\begin{figure*}[t]
  \centering
   \includegraphics[width=0.9\linewidth]{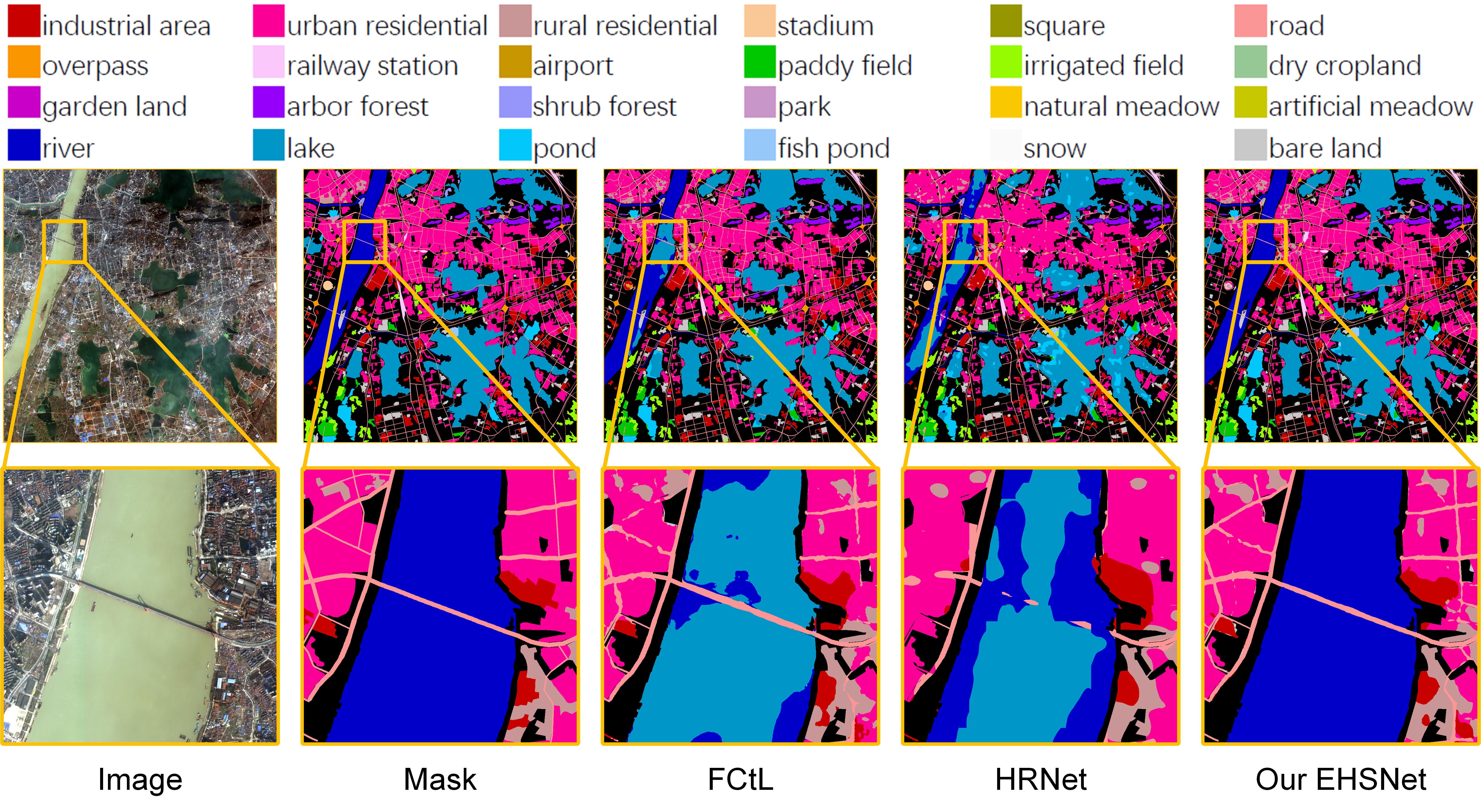}
   \caption{\textbf{We illustrate some representative examples of the FBP dataset.} The competitor methods falsely predict the river into the lake, while our EHSNet shows superiority in preserving complete object contours, as demonstrated in the {\color{myorange}orange} boxes.}
   \label{fig:6}
\end{figure*}

\subsection{Implementation details}

We implement our framework and all the comparison methods using Pytorch on a server with three NVIDIA RTX Titan GPU, each of which has 24G GPU memory. The server has 1024G host memory. 

Considering both efficacy and effectiveness, we choose HRNet \cite{hrnet} as the backbone model to build our EHSNet model. HRNet is a powerful and efficient model which can obtain multi-scale features. It is noted that we change the output stride from 4 to 8 to meet the requirements of GPU memories. For the parameters. The dimensions of features $d$ and $k$ in the LRD module are both set to 256. The percentage for hard region selection $a$ in the ECR module is set to 10\%. The kernel size of the Sobel operator in the BAE module is set to 5. The weight of the auxiliary segmentation loss is 0.4 following the routine \cite{pspnet}. We use the mean intersection over union (mIoU) as the evaluation metric.

During training our local segmentation model, we adopt the SGD optimizer and a mini-batch size of 3 on the FBP dataset and 6 on the Inria Aerial dataset. The initial learning rate is set to 0.01 and it is decayed to 0.0001 by a step-based learning rate policy. In practice, it takes 10,000 iterations to converge our EHSNet model. For the first 1,500 iterations, the learning rate is set to 5e-5 for warming up. Following \cite{kirillov2020pointrend}, we multiply the learning rate for the segmentation heads by 10 for better convergence.

\begin{table}
  \centering
  \begin{tabular}{c |c | c}
    \toprule
     Methods & Type & mIoU (\%) \\
    \midrule
    GoogLeNet \cite{szegedy2015google}  &\multirow{9}{*}{Generic SS}      & 28.99 \\
    ResNet \cite{he2016resnet}     &      & 33.59 \\
    UNet \cite{ronneberger2015unet}       &      & 44.51 \\
    PSPNet \cite{ronneberger2015unet}     &      & 53.33 \\
    DeepLabv3+ \cite{deeplabv3p}  &      & 48.97 \\
    UperNet \cite{xiao2018upernet}    &      & 51.08 \\
    HRNet \cite{hrnet}      &      & 53.83 \\
    STDC \cite{stdc}       &      & 51.29 \\
    Swin \cite{swin}       &      & 50.49 \\
    \midrule
    GLNet \cite{glnet}      &\multirow{4}{*}{UHRIS (LRISS)}       & 42.05 \\
    FCtL \cite{fctl}       &      & 48.28 \\
    MagNet \cite{magnet}     &      & 44.20 \\
    ISDNet \cite{guo2022isdnet}     &      & 21.98 \\
    \midrule
    EHSNet    &    LRISS    & \textbf{59.48(+5.65)} \\
    \bottomrule
    \end{tabular}
    \vspace{0.8em}
    \caption{\textbf{Comparison with previous SOTAs on the FBP dataset.} We evaluate several typical generic SS methods and LRISS methods on the FBP dataset. The codes released by \cite{glnet, fctl, magnet, guo2022isdnet} are modified to adjust to the FBP dataset.}
    \label{tab:fbp}
\end{table}

\begin{table}
  \centering
  \begin{tabular}{c | c|c}
    \toprule
     Methods & Type & mIoU (\%) \\
    \midrule
    ICNet \cite{zhao2018icnet}        &\multirow{4}{*}{Generic SS}    &  31.10   \\
    FCN \cite{fcn}          &    &  69.10   \\
    DeepLabv3+ \cite{deeplabv3p}   &    &  55.90   \\
    STDC \cite{stdc}         &    &  72.44   \\
    \midrule
    GLNet \cite{glnet}        &\multirow{4}{*}{UHRIS (LRISS)}    &  71.20   \\
    CascadePSP \cite{caspsp}   &    &  69.40   \\
    FCtL \cite{fctl}         &    &  72.87   \\
    ISDNet \cite{guo2022isdnet}       &    &  74.23   \\
    \midrule
    EHSNet       & LRISS    & \textbf{78.51(+4.28)}   \\
    \bottomrule
    \end{tabular}
    \vspace{0.8em}
    \caption{\textbf{Comparison with previous SOTAs on the Inria Aerial dataset.} The results of existing methods on the Inria Aerial dataset are collected from \cite{glnet, fctl, guo2022isdnet}.}
    \label{tab:inria}
\end{table}

\subsection{Comparison with state-of-the-arts}
The sizes of images in the FBP dataset and the Inria Aerial dataset are both large. The Inria Aerial dataset only contains one class of object which is building, and the annotations are coarse, which makes the LRISS much easier. While the FBP dataset is much harder compared to the Inria Aerial dataset, containing 24 classes. 

\begin{figure*}[h]
  \centering
   \includegraphics[width=0.9\linewidth]{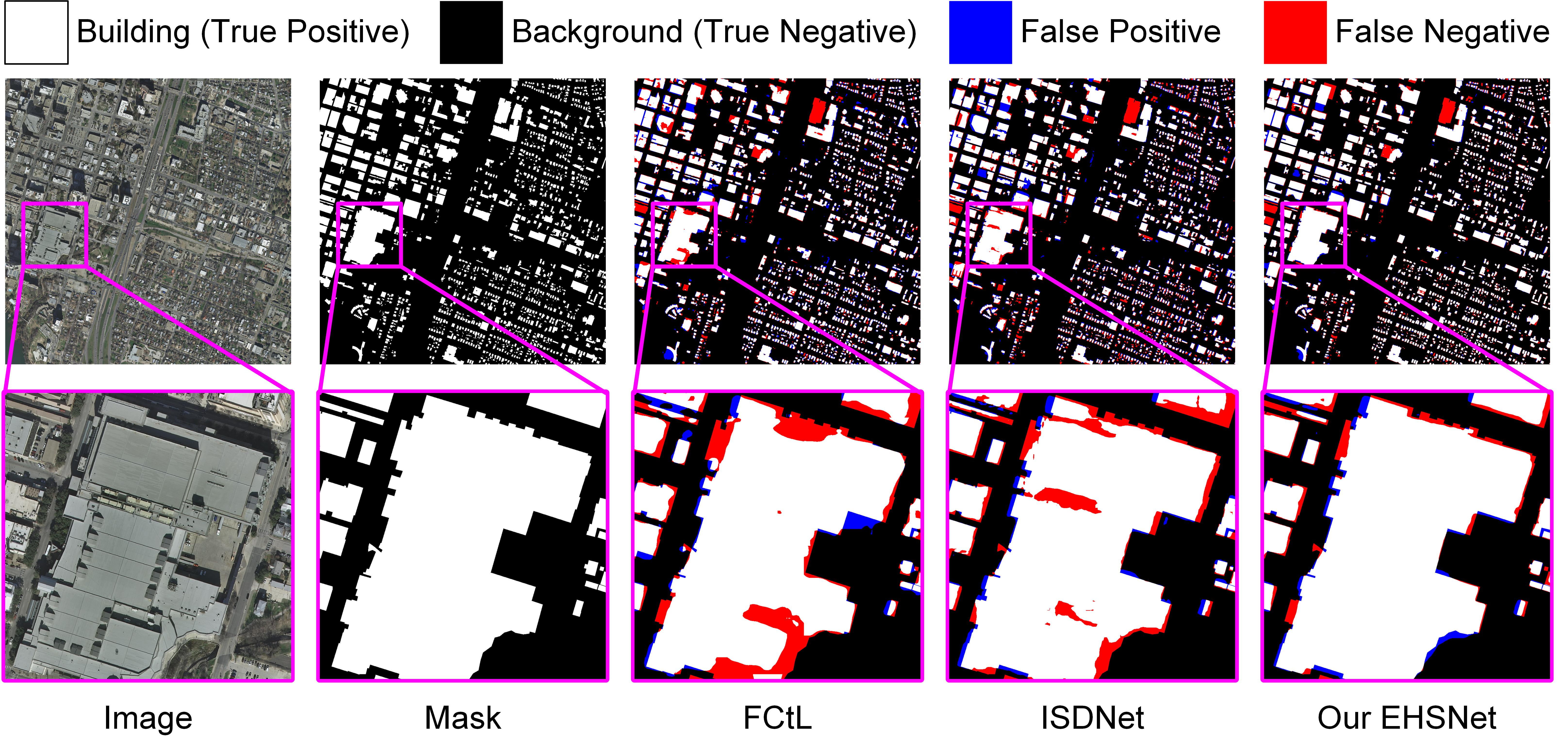}
   \caption{\textbf{We illustrate some representative examples of the Inria Aerial dataset.} Compared with previous SOTAs, our proposed EHSNet can preserve complete building objects, with much fewer commission errors ({\color{blue}{false positive}}) and omission errors ({\color{red}{false negative}}), as demonstrated in the {\color{mypink}pink} boxes.}
   \label{fig:7}
\end{figure*}
\begin{table*}
  \centering
  \begin{tabular}{c | c | c c | c c}
    \toprule
     Methods & mIoU (\%) & GPU memory (GB) & Host memory (GB) & Training time (sec) & Testing time (sec) \\
    \midrule
    GLNet \cite{glnet}            & 42.05 & \multirow{4}{*}{24}  & 285           & 344 & 54 \\
    FCtL \cite{fctl}             & 48.28 &                      & 164           & 356 & 24 \\
    MagNet \cite{magnet}           & 44.20 &                      & \textbf{4.5}  & 3.2 & 34 \\
    ISDNet \cite{guo2022isdnet}    & 21.98 &      & 10.1          & \textbf{1.3} & \textbf{2.8} \\
    \midrule
    EHSNet           & \textbf{59.48} & 24 & 117 & 16.6 & 4.8\\
    \bottomrule
    \end{tabular}
    \vspace{0.8em}
    \caption{\textbf{Efficiency comparison with previous LRISS methods on the FBP dataset.} Our EHSNet can achieve great performance gains with fewer resources and less time needed than both GLNet \cite{glnet} and FCtL \cite{fctl}. Although ISDNet \cite{guo2022isdnet} is the lightest and fastest, it does not perform well on the FBP dataset. The memory cost is calculated with the maximum mini-batch sizes of each during training. And the time cost is calculated with the forwarding procedure of one LRI.}
    \label{tab:efficiency}
\end{table*}

\textbf{FBP \cite{fbp}.} 
We collect the results of GoogLeNet, ResNet, and UNet from \cite{fbp}. The other results on the FBP dataset are re-implemented by us based on the MMsegmentation \cite{mmseg2020} framework. Specifically, six typical generic SS methods \cite{pspnet, deeplabv3p, xiao2018upernet, hrnet, stdc, swin} are selected, including one popular transformer-based method, the Swin Transformer \cite{swin}. And we re-implement four typical UHRIS (LRISS) methods \cite{glnet, fctl, magnet, guo2022isdnet}. It is noted that we do not include CascadePSP \cite{caspsp} and HQS \cite{shen2022highquality} for comparison since they both need pre-trained models to predict initial segmentation for refinement.

As can be seen in \cref{tab:fbp}, existing LRISS methods fail to work well on the FBP dataset since they can not fully utilize the rich information in the LRIs, while our EHSNet performs well with holistic learning. EHSNet outperforms the SOTA method by a significant margin of +5.65 mIoU. As shown in \cref{fig:6}, the competitor methods falsely predict the river into the lake. However, our proposed EHSNet can utilize complete information with holistic learning and combine the advantage of global and local views to obtain an accurate segmentation result.

\textbf{Inria Aerial \cite{inria}.}
 The competitor results on the Inria Aerial dataset are collected from previous SOTA works \cite{glnet, fctl, guo2022isdnet}. As can be seen in \cref{tab:inria}, EHSNet outperforms the SOTA method by a large margin of +4.28 mIoU. Some qualitative results are presented in \cref{fig:7}, which also demonstrate that our proposed EHSNet is superior to the previous SOTA methods. The competitor methods \cite{fctl, guo2022isdnet} falsely segment the whole building into separated ones with incomplete boundaries. However, our proposed EHSNet can preserve complete building objects, with much fewer commission errors (false positive) and omission errors (false negative), as shown in \cref{fig:7}.

    \begin{table*}
    \centering
    \begin{tabular}{c|cccc|cc}
    \toprule
    \quad & Holistic learning & BAE & LRD & ECR & mIoU (\%)  & $\Delta$ (\%) \\
    \midrule
    Baseline  & -           & -          & -      & -           & 45.93  & -\\
    I    & \checkmark  & -          & -          & -          & 48.67 &\textcolor{blue}{$+$2.74} \\
    II   & \checkmark  & \checkmark & -          & -          & 49.49 &\textcolor{blue}{$+$3.56} \\
    III  & \checkmark  & -          & \checkmark & -          & 49.61 &\textcolor{blue}{$+$3.68} \\
    IV   & \checkmark  & -          & -          & \checkmark & 49.66 &\textcolor{blue}{$+$3.73} \\
    V    & \checkmark  & \checkmark & \checkmark & -          & 50.55 &\textcolor{blue}{$+$4.62} \\
    VI   & \checkmark  & -          & \checkmark & \checkmark & 51.00 &\textcolor{blue}{$+$5.07} \\
    VII  & \checkmark  & \checkmark & -          & \checkmark & 51.10 &\textcolor{blue}{$+$5.17} \\
      \midrule
    EHSNet  & \checkmark  & \checkmark & \checkmark & \checkmark & \textbf{52.42} &\textbf{\textcolor{blue}{$+$6.49}} \\
      \bottomrule
    \end{tabular}
    \vspace{0.8em}
    \caption{\textbf{Ablation study of proposed modules on the FBP dataset.} It shows the specific effectiveness of the three proposed modules on the setting of holistic learning.}
    \label{tab:fbpablation}
\end{table*}

\begin{table}
  \centering
  \begin{tabular}{c | cc}
    \toprule
     Methods & $a$ & mIoU (\%)  \\
    \midrule
    I   & 1.25  & 42.03  \\
    II  & 2.5  & 43.32  \\
    III & 5  & 43.51   \\
    IV  & 10  & \textbf{43.62}  \\
    V   & 20  & 42.44   \\
    \bottomrule
    \end{tabular}
    \vspace{0.8em}
    \caption{\textbf{Sensitivity analysis} with the percentage for hard region selection $a$ in the ECR module on the FBP dataset.}
    \label{tab:param1}
\end{table}

\begin{table}
  \centering
  \begin{tabular}{c | cc}
    \toprule
     Methods & $k$ & mIoU (\%) \\
    \midrule
    I   & 128  & 42.54 \\
    II  & 256  & \textbf{43.62} \\
    III  & 384  & 43.48 \\
    IV    & 512  & 43.33 \\
    \bottomrule
    \end{tabular}
    \vspace{0.8em}
     \caption{\textbf{Sensitivity analysis} with the dimensions of features $k$ in the LRD module on the FBP dataset.}
    \label{tab:param2}
\end{table}

\subsection{Efficiency}
We evaluate the efficiencies of four typical LRISS methods \cite{glnet, fctl, magnet, guo2022isdnet} and our EHSNet. The experiments are all conducted with 24G GPU memory and 1024G host memory. We adjust the mini-batch size to be as large as possible, to make full use of GPU memory for a fair comparison. The cost of host memory is calculated during all the training stages of each model. The time cost is calculated with the forwarding procedure of one LRI.

As can be seen from \cref{tab:efficiency}, the time and memory cost of GLNet \cite{glnet} and FCtL \cite{fctl} is too much. It takes more than a week for training one stage and about a month for the whole training process with much memory occupied. Although MagNet \cite{magnet} and ISDNet \cite{guo2022isdnet} cost less time and memory, they fail to achieve desirable performances. The original version of MagNet \cite{magnet} includes four-stage refinement. But we can only re-implement one-stage refinement according to their released code. Besides, the model of ISDNet \cite{guo2022isdnet} is specially designed for 3-band images, and we have to modify it to fit the 4-band images on the FBP dataset but its result is not promising. 

\subsection{Ablation study}
We ablate the proposed three modules and the holistic learning to see whether they are really effective for LRISS. The results are in \cref{tab:fbpablation}. It turns out that holistic learning can bring a significant performance gain with +2.74 mIoU. Based on that, our proposed three modules including BAE, LRD, and ECR can help to further enhance the performance with +0.82, +0.94, +0.99 mIoU, respectively. With all proposed modules together, the performance gain is +6.49 mIoU in total.

\subsection{Sensitivity analysis}
We analyze the sensitivity of parameters in our proposed modules with a light version of EHSNet. The results in \cref{tab:param1} and \cref{tab:param2} illustrate that the best results can be obtained when $a$ and $k$ are set to 10 and 256, respectively.

\section{Conclusion and limitation}
\label{sec:conclucion}
\subsection{Conclusion}
    This paper presents a new scheme named EHSNet which can capably process holistic LRISS and overcome the GPU memory exhaustion caused by the extremely large sizes of LRIs. Different from existing global-local fusion methods or multi-stage refinement methods, EHSNet can elegantly process holistic LRISS and is able to thoroughly exploit the abundant information in LRIs. By extending the tensor storage from GPU memory to host memory, EHSNet breaks through the constraint of limited GPU memory. To the best of our knowledge, it is the first time that holistic LRISS can be done. Besides, three modules are specifically designed to utilize the characteristics of LRIs, including the long-range dependency module, the efficient cross-correlation module, and the boundary-aware enhancement module. Extensive experiments show that EHSNet outperforms the previous SOTAs on two typical LRISS datasets. We hope that EHSNet can open a new perspective for LRISS.
    
\subsection{Limitation}
\textbf{Optimization.} The mini-batch size during the training of EHSNet is quite small while the image size is extremely large under the setting of LRISS. We believe that new optimization methods need to be explored for LRISS. 

\textbf{Efficiency.} For now, EHSNet is not efficient enough. Some efficient techniques for accelerating the process of LRISS are supposed to be further developed, which are not limited to the attention function.


{\small
\bibliographystyle{ieee_fullname}
\bibliography{egbib}

\begin{thebibliography}{10}\itemsep=-1pt

\bibitem{boundary1}
Gedas Bertasius, Jianbo Shi, and Lorenzo Torresani.
\newblock Semantic segmentation with boundary neural fields.
\newblock In {\em Proceedings of the IEEE conference on computer vision and
  pattern recognition}, pages 3602--3610, 2016.

\bibitem{gpt}
Tom Brown, Benjamin Mann, Nick Ryder, Melanie Subbiah, Jared~D Kaplan, Prafulla
  Dhariwal, Arvind Neelakantan, Pranav Shyam, Girish Sastry, Amanda Askell,
  et~al.
\newblock Language models are few-shot learners.
\newblock {\em Advances in neural information processing systems},
  33:1877--1901, 2020.

\bibitem{castillo2021semi}
Javiera Castillo-Navarro, Bertrand Le~Saux, Alexandre Boulch, Nicolas Audebert,
  and S{\'e}bastien Lef{\`e}vre.
\newblock Semi-supervised semantic segmentation in earth observation: The
  minifrance suite, dataset analysis and multi-task network study.
\newblock {\em Machine Learning}, pages 1--36, 2021.

\bibitem{deeplabv3p}
Liang-Chieh Chen, Yukun Zhu, George Papandreou, Florian Schroff, and Hartwig
  Adam.
\newblock Encoder-decoder with atrous separable convolution for semantic image
  segmentation.
\newblock In {\em ECCV}, pages 801--818, 2018.

\bibitem{glnet}
Wuyang Chen, Ziyu Jiang, Zhangyang Wang, Kexin Cui, and Xiaoning Qian.
\newblock Collaborative global-local networks for memory-efficient segmentation
  of ultra-high resolution images.
\newblock In {\em CVPR}, pages 8924--8933, 2019.

\bibitem{caspsp}
Ho~Kei Cheng, Jihoon Chung, Yu-Wing Tai, and Chi-Keung Tang.
\newblock Cascadepsp: toward class-agnostic and very high-resolution
  segmentation via global and local refinement.
\newblock In {\em CVPR}, pages 8890--8899, 2020.

\bibitem{mmseg2020}
MMSegmentation Contributors.
\newblock {MMSegmentation}: Openmmlab semantic segmentation toolbox and
  benchmark.
\newblock \url{https://github.com/open-mmlab/mmsegmentation}, 2020.

\bibitem{dai2017deformable}
Jifeng Dai, Haozhi Qi, Yuwen Xiong, Yi Li, Guodong Zhang, Han Hu, and Yichen
  Wei.
\newblock Deformable convolutional networks.
\newblock In {\em Proceedings of the IEEE international conference on computer
  vision}, pages 764--773, 2017.

\bibitem{humanvision2}
Mihai Datcu and Klaus Seidel.
\newblock Human-centered concepts for exploration and understanding of earth
  observation images.
\newblock {\em IEEE Transactions on Geoscience and Remote Sensing},
  43(3):601--609, 2005.

\bibitem{wiconet}
Lei Ding, Dong Lin, Shaofu Lin, Jing Zhang, Xiaojie Cui, Yuebin Wang, Hao Tang,
  and Lorenzo Bruzzone.
\newblock Looking outside the window: Wide-context transformer for the semantic
  segmentation of high-resolution remote sensing images.
\newblock {\em IEEE Transactions on Geoscience and Remote Sensing}, 60:1--13,
  2022.

\bibitem{tsmta}
Lei Ding, Jing Zhang, and Lorenzo Bruzzone.
\newblock Semantic segmentation of large-size vhr remote sensing images using a
  two-stage multiscale training architecture.
\newblock {\em IEEE Transactions on Geoscience and Remote Sensing},
  58(8):5367--5376, 2020.

\bibitem{ding2022replknet}
Xiaohan Ding, Xiangyu Zhang, Jungong Han, and Guiguang Ding.
\newblock Scaling up your kernels to 31x31: Revisiting large kernel design in
  cnns.
\newblock In {\em Proceedings of the IEEE/CVF Conference on Computer Vision and
  Pattern Recognition}, pages 11963--11975, 2022.

\bibitem{vit}
Alexey Dosovitskiy, Lucas Beyer, Alexander Kolesnikov, Dirk Weissenborn,
  Xiaohua Zhai, Thomas Unterthiner, Mostafa Dehghani, Matthias Minderer, Georg
  Heigold, Sylvain Gelly, et~al.
\newblock An image is worth 16x16 words: Transformers for image recognition at
  scale.
\newblock In {\em ICLR}, 2021.

\bibitem{stdc}
Mingyuan Fan, Shenqi Lai, Junshi Huang, Xiaoming Wei, Zhenhua Chai, Junfeng
  Luo, and Xiaolin Wei.
\newblock Rethinking bisenet for real-time semantic segmentation.
\newblock In {\em Proceedings of the IEEE/CVF conference on computer vision and
  pattern recognition}, pages 9716--9725, 2021.

\bibitem{gongpeng}
Peng Gong, Han Liu, Meinan Zhang, Congcong Li, Jie Wang, Huabing Huang,
  Nicholas Clinton, Luyan Ji, Wenyu Li, Yuqi Bai, Bin Chen, Bing Xu, Zhiliang
  Zhu, Cui Yuan, Hoi {Ping Suen}, Jing Guo, Nan Xu, Weijia Li, Yuanyuan Zhao,
  Jun Yang, Chaoqing Yu, Xi Wang, Haohuan Fu, Le Yu, Iryna Dronova, Fengming
  Hui, Xiao Cheng, Xueli Shi, Fengjin Xiao, Qiufeng Liu, and Lianchun Song.
\newblock Stable classification with limited sample: transferring a 30-m
  resolution sample set collected in 2015 to mapping 10-m resolution global
  land cover in 2017.
\newblock {\em Science Bulletin}, 64(6):370--373, 2019.

\bibitem{guo2022isdnet}
Shaohua Guo, Liang Liu, Zhenye Gan, Yabiao Wang, Wuhao Zhang, Chengjie Wang,
  Guannan Jiang, Wei Zhang, Ran Yi, Lizhuang Ma, et~al.
\newblock Isdnet: Integrating shallow and deep networks for efficient
  ultra-high resolution segmentation.
\newblock In {\em Proceedings of the IEEE/CVF Conference on Computer Vision and
  Pattern Recognition}, pages 4361--4370, 2022.

\bibitem{humanvision1}
Bart M~Ter Haar~Romeny and Luc Florack.
\newblock A multiscale geometric model of human vision.
\newblock In {\em The Perception of Visual Information}, pages 73--114.
  Springer, 1993.

\bibitem{he2016resnet}
Kaiming He, Xiangyu Zhang, Shaoqing Ren, and Jian Sun.
\newblock Deep residual learning for image recognition.
\newblock In {\em Proceedings of the IEEE conference on computer vision and
  pattern recognition}, pages 770--778, 2016.

\bibitem{magnet}
Chuong Huynh, Anh~Tuan Tran, Khoa Luu, and Minh Hoai.
\newblock Progressive semantic segmentation.
\newblock In {\em Proceedings of the IEEE/CVF Conference on Computer Vision and
  Pattern Recognition}, pages 16755--16764, 2021.

\bibitem{jiang2020byteps}
Yimin Jiang, Yibo Zhu, Chang Lan, Bairen Yi, Yong Cui, and Chuanxiong Guo.
\newblock A unified architecture for accelerating distributed dnn training in
  heterogeneous gpu/cpu clusters.
\newblock In {\em Proceedings of the 14th USENIX Conference on Operating
  Systems Design and Implementation}, pages 463--479, 2020.

\bibitem{kirillov2020pointrend}
Alexander Kirillov, Yuxin Wu, Kaiming He, and Ross Girshick.
\newblock Pointrend: Image segmentation as rendering.
\newblock In {\em Proceedings of the IEEE/CVF conference on computer vision and
  pattern recognition}, pages 9799--9808, 2020.

\bibitem{emergency2019}
Christos Kyrkou and Theocharis Theocharides.
\newblock Deep-learning-based aerial image classification for emergency
  response applications using unmanned aerial vehicles.
\newblock In {\em CVPR workshops}, pages 517--525, 2019.

\bibitem{lms}
Tung~D. Le, Haruki Imai, Yasushi Negishi, and Kiyokuni Kawachiya.
\newblock Automatic gpu memory management for large neural models in
  tensorflow.
\newblock In {\em Proceedings of the 2019 ACM SIGPLAN International Symposium
  on Memory Management}, ISMM 2019, page 1–13, New York, NY, USA, 2019.
  Association for Computing Machinery.

\bibitem{lee2015deepsupervision}
Chen-Yu Lee, Saining Xie, Patrick Gallagher, Zhengyou Zhang, and Zhuowen Tu.
\newblock Deeply-supervised nets.
\newblock In {\em Artificial intelligence and statistics}, pages 562--570.
  PMLR, 2015.

\bibitem{fctl}
Qi Li, Weixiang Yang, Wenxi Liu, Yuanlong Yu, and Shengfeng He.
\newblock From contexts to locality: Ultra-high resolution image segmentation
  via locality-aware contextual correlation.
\newblock In {\em ICCV}, pages 7252--7261, 2021.

\bibitem{building2021}
Weijia Li, Wenqian Zhao, Huaping Zhong, Conghui He, and Dahua Lin.
\newblock Joint semantic-geometric learning for polygonal building
  segmentation.
\newblock In {\em Proceedings of the AAAI Conference on Artificial
  Intelligence}, volume~35, pages 1958--1965, 2021.

\bibitem{li2017notallrefine}
Xiaoxiao Li, Ziwei Liu, Ping Luo, Chen Change~Loy, and Xiaoou Tang.
\newblock Not all pixels are equal: Difficulty-aware semantic segmentation via
  deep layer cascade.
\newblock In {\em Proceedings of the IEEE conference on computer vision and
  pattern recognition}, pages 3193--3202, 2017.

\bibitem{mfvnet}
Yansheng Li, Wei Chen, Xin Huang, Zhi Gao, Siwei Li, Tao He, and Zhang Yongjun.
\newblock Mfvnet: Deep adaptive fusion network with multiple field-of-views for
  remote sensing image semantic segmentation.
\newblock {\em SCIENCE CHINA Information Sciences}, 2022.

\bibitem{lin2017refinenet}
Guosheng Lin, Anton Milan, Chunhua Shen, and Ian Reid.
\newblock Refinenet: Multi-path refinement networks for high-resolution
  semantic segmentation.
\newblock In {\em Proceedings of the IEEE conference on computer vision and
  pattern recognition}, pages 1925--1934, 2017.

\bibitem{liu2022swin}
Ze Liu, Han Hu, Yutong Lin, Zhuliang Yao, Zhenda Xie, Yixuan Wei, Jia Ning, Yue
  Cao, Zheng Zhang, Li Dong, et~al.
\newblock Swin transformer v2: Scaling up capacity and resolution.
\newblock In {\em Proceedings of the IEEE/CVF Conference on Computer Vision and
  Pattern Recognition}, pages 12009--12019, 2022.

\bibitem{swin}
Ze Liu, Yutong Lin, Yue Cao, Han Hu, Yixuan Wei, Zheng Zhang, Stephen Lin, and
  Baining Guo.
\newblock Swin transformer: Hierarchical vision transformer using shifted
  windows.
\newblock In {\em Proceedings of the IEEE/CVF International Conference on
  Computer Vision (ICCV)}, pages 10012--10022, October 2021.

\bibitem{fcn}
Jonathan Long, Evan Shelhamer, and Trevor Darrell.
\newblock Fully convolutional networks for semantic segmentation.
\newblock In {\em CVPR}, pages 3431--3440, 2015.

\bibitem{inria}
Emmanuel Maggiori, Yuliya Tarabalka, Guillaume Charpiat, and Pierre Alliez.
\newblock Can semantic labeling methods generalize to any city? the inria
  aerial image labeling benchmark.
\newblock In {\em 2017 IEEE International Geoscience and Remote Sensing
  Symposium (IGARSS)}, pages 3226--3229, 2017.

\bibitem{humanvision}
Juhong Min, Yucheng Zhao, Chong Luo, and Minsu Cho.
\newblock {Peripheral Vision Transformer}.
\newblock In {\em Advances in Neural Information Processing Systems}, 2022.

\bibitem{peng2017largekernelmatters}
Chao Peng, Xiangyu Zhang, Gang Yu, Guiming Luo, and Jian Sun.
\newblock Large kernel matters--improve semantic segmentation by global
  convolutional network.
\newblock In {\em Proceedings of the IEEE conference on computer vision and
  pattern recognition}, pages 4353--4361, 2017.

\bibitem{rasley2020deepspeed}
Jeff Rasley, Samyam Rajbhandari, Olatunji Ruwase, and Yuxiong He.
\newblock Deepspeed: System optimizations enable training deep learning models
  with over 100 billion parameters.
\newblock In {\em Proceedings of the 26th ACM SIGKDD International Conference
  on Knowledge Discovery \& Data Mining}, pages 3505--3506, 2020.

\bibitem{rsapp1}
Alexandre Robicquet, Amir Sadeghian, Alexandre Alahi, and Silvio Savarese.
\newblock Learning social etiquette: Human trajectory understanding in crowded
  scenes.
\newblock In {\em European conference on computer vision}, pages 549--565.
  Springer, 2016.

\bibitem{ronneberger2015unet}
Olaf Ronneberger, Philipp Fischer, and Thomas Brox.
\newblock U-net: Convolutional networks for biomedical image segmentation.
\newblock In {\em Medical Image Computing and Computer-Assisted
  Intervention--MICCAI 2015: 18th International Conference, Munich, Germany,
  October 5-9, 2015, Proceedings, Part III 18}, pages 234--241. Springer, 2015.

\bibitem{shen2022highquality}
Tiancheng Shen, Yuechen Zhang, Lu Qi, Jason Kuen, Xingyu Xie, Jianlong Wu, Zhe
  Lin, and Jiaya Jia.
\newblock High quality segmentation for ultra high-resolution images.
\newblock In {\em Proceedings of the IEEE/CVF Conference on Computer Vision and
  Pattern Recognition}, pages 1310--1319, 2022.

\bibitem{shen2021efficientattn}
Zhuoran Shen, Mingyuan Zhang, Haiyu Zhao, Shuai Yi, and Hongsheng Li.
\newblock Efficient attention: Attention with linear complexities.
\newblock In {\em Proceedings of the IEEE/CVF winter conference on applications
  of computer vision}, pages 3531--3539, 2021.

\bibitem{strudel2021segmenter}
Robin Strudel, Ricardo Garcia, Ivan Laptev, and Cordelia Schmid.
\newblock Segmenter: Transformer for semantic segmentation.
\newblock In {\em Proceedings of the IEEE/CVF International Conference on
  Computer Vision}, pages 7262--7272, 2021.

\bibitem{szegedy2015google}
Christian Szegedy, Wei Liu, Yangqing Jia, Pierre Sermanet, Scott Reed, Dragomir
  Anguelov, Dumitru Erhan, Vincent Vanhoucke, and Andrew Rabinovich.
\newblock Going deeper with convolutions.
\newblock In {\em Proceedings of the IEEE conference on computer vision and
  pattern recognition}, pages 1--9, 2015.

\bibitem{boundary2}
Towaki Takikawa, David Acuna, Varun Jampani, and Sanja Fidler.
\newblock Gated-scnn: Gated shape cnns for semantic segmentation.
\newblock In {\em Proceedings of the IEEE/CVF international conference on
  computer vision}, pages 5229--5238, 2019.

\bibitem{fbp}
Xin-Yi Tong, Gui-Song Xia, Qikai Lu, Huanfeng Shen, Shengyang Li, Shucheng You,
  and Liangpei Zhang.
\newblock Land-cover classification with high-resolution remote sensing images
  using transferable deep models.
\newblock {\em Remote Sensing of Environment}, 237:111322, 2020.

\bibitem{hrnet}
Jingdong Wang, Ke Sun, Tianheng Cheng, Borui Jiang, Chaorui Deng, Yang Zhao,
  Dong Liu, Yadong Mu, Mingkui Tan, Xinggang Wang, et~al.
\newblock Deep high-resolution representation learning for visual recognition.
\newblock {\em PAMI}, 2020.

\bibitem{wang2018nonlocal}
Xiaolong Wang, Ross Girshick, Abhinav Gupta, and Kaiming He.
\newblock Non-local neural networks.
\newblock In {\em Proceedings of the IEEE conference on computer vision and
  pattern recognition}, pages 7794--7803, 2018.

\bibitem{wu2020ppn}
Tong Wu, Zhenzhen Lei, Bingqian Lin, Cuihua Li, Yanyun Qu, and Yuan Xie.
\newblock Patch proposal network for fast semantic segmentation of
  high-resolution images.
\newblock In {\em Proceedings of the AAAI Conference on Artificial
  Intelligence}, volume~34, pages 12402--12409, 2020.

\bibitem{xiao2018upernet}
Tete Xiao, Yingcheng Liu, Bolei Zhou, Yuning Jiang, and Jian Sun.
\newblock Unified perceptual parsing for scene understanding.
\newblock In {\em Proceedings of the European conference on computer vision
  (ECCV)}, pages 418--434, 2018.

\bibitem{segformer}
Enze Xie, Wenhai Wang, Zhiding Yu, Anima Anandkumar, Jose~M. Alvarez, and Ping
  Luo.
\newblock Segformer: Simple and efficient design for semantic segmentation with
  transformers.
\newblock In M. Ranzato, A. Beygelzimer, Y. Dauphin, P.S. Liang, and J.~Wortman
  Vaughan, editors, {\em Advances in Neural Information Processing Systems},
  volume~34, pages 12077--12090. Curran Associates, Inc., 2021.

\bibitem{yuan2020ocrnet}
Yuhui Yuan, Xilin Chen, and Jingdong Wang.
\newblock Object-contextual representations for semantic segmentation.
\newblock In {\em European conference on computer vision}, pages 173--190.
  Springer, 2020.

\bibitem{yuan2020segfix}
Yuhui Yuan, Jingyi Xie, Xilin Chen, and Jingdong Wang.
\newblock Segfix: Model-agnostic boundary refinement for segmentation.
\newblock In {\em European Conference on Computer Vision}, pages 489--506.
  Springer, 2020.

\bibitem{zhao2020selfattn}
Hengshuang Zhao, Jiaya Jia, and Vladlen Koltun.
\newblock Exploring self-attention for image recognition.
\newblock In {\em Proceedings of the IEEE/CVF Conference on Computer Vision and
  Pattern Recognition}, pages 10076--10085, 2020.

\bibitem{zhao2018icnet}
Hengshuang Zhao, Xiaojuan Qi, Xiaoyong Shen, Jianping Shi, and Jiaya Jia.
\newblock Icnet for real-time semantic segmentation on high-resolution images.
\newblock In {\em Proceedings of the European conference on computer vision
  (ECCV)}, pages 405--420, 2018.

\bibitem{pspnet}
Hengshuang Zhao, Jianping Shi, Xiaojuan Qi, Xiaogang Wang, and Jiaya Jia.
\newblock Pyramid scene parsing network.
\newblock In {\em CVPR}, pages 2881--2890, 2017.

\bibitem{zhao2018psanet}
Hengshuang Zhao, Yi Zhang, Shu Liu, Jianping Shi, Chen~Change Loy, Dahua Lin,
  and Jiaya Jia.
\newblock Psanet: Point-wise spatial attention network for scene parsing.
\newblock In {\em Proceedings of the European conference on computer vision
  (ECCV)}, pages 267--283, 2018.

\bibitem{setr}
Sixiao Zheng, Jiachen Lu, Hengshuang Zhao, Xiatian Zhu, Zekun Luo, Yabiao Wang,
  Yanwei Fu, Jianfeng Feng, Tao Xiang, Philip~HS Torr, et~al.
\newblock Rethinking semantic segmentation from a sequence-to-sequence
  perspective with transformers.
\newblock In {\em CVPR}, pages 6881--6890, 2021.

\end{thebibliography}
}

\end{document}